\documentclass[11pt,a4paper]{article}
\usepackage[dvipdfmx]{graphicx}
\usepackage[hyperref]{emnlp2020}
\usepackage{times}
\usepackage{latexsym}

\usepackage{bm}
\usepackage{booktabs}
\usepackage{amsmath}
\usepackage{amsfonts}
\usepackage{multirow}
\usepackage{array}
\newcolumntype{L}[1]{>{\raggedright\let\newline\\\arraybackslash\hspace{0pt}}m{#1}}
\newcolumntype{C}[1]{>{\centering\let\newline\\\arraybackslash\hspace{0pt}}m{#1}}
\newcolumntype{R}[1]{>{\raggedleft\let\newline\\\arraybackslash\hspace{0pt}}m{#1}}

\usepackage{chngcntr}
\usepackage{afterpage}
\usepackage{float}

\usepackage{microtype}

\aclfinalcopy %

\title{Unsupervised Keyword Extraction for Full-Sentence VQA}

\author{Kohei Uehara \\
The University of Tokyo \\
\texttt{uehara@mi.t.u-tokyo.ac.jp} \\\And
Tatsuya Harada \\
The University of Tokyo \\
RIKEN \\
\texttt{harada@mi.t.u-tokyo.ac.jp} \\}

\date{}

\begin{document}
\maketitle
\begin{abstract}
In the majority of the existing Visual Question Answering (VQA) research, the answers consist of short, often single words, as per instructions given to the annotators during dataset construction.
This study envisions a VQA task for natural situations, where the answers are more likely to be sentences rather than single words.
To bridge the gap between this natural VQA and existing VQA approaches, a novel unsupervised keyword extraction method is proposed.
The method is based on the principle that the full-sentence answers can be decomposed into two parts: one that contains new information answering the question (i.e., keywords), and one that contains information already included in the question.
Discriminative decoders were designed to achieve such decomposition, and the method was experimentally implemented on VQA datasets containing full-sentence answers.
The results show that the proposed model can accurately extract the keywords without being given explicit annotations describing them.
\end{abstract}

\section{Introduction}\label{intro}
Visual recognition is one of the most actively researched fields; this research is expected to be applied to real-world systems such as robots.
Since innumerable object classes exist in the real world, training all of them in advance is impossible.
Thus, to train image recognition models, it is important for real-world intelligent systems to actively acquire information.
One promising approach to acquire information on the fly is \textit{learning by asking}, i.e., generating questions to humans about unknown objects, and consequently learning new knowledge from the human response~\citep{lba, vqg_unknown,vqg_caption}.
This implies that if we can build a Visual Question Answering (VQA) system~\citep{vqa} that functions in the real world and extracts knowledge from human responses, we can realize an intelligent system that can learn autonomously.

VQA is a well-known vision and language task which aims to develop a system that can answer a question about an image.
One typical dataset used in VQA is the VQA v2 dataset~\citep{vqav2}.
The answers in the VQA v2 dataset are essentially single words.
This is because the annotators are instructed to keep the answer as short as possible when constructing the dataset.

\begin{figure}[t]
\begin{center}
\includegraphics[width=0.8\linewidth]{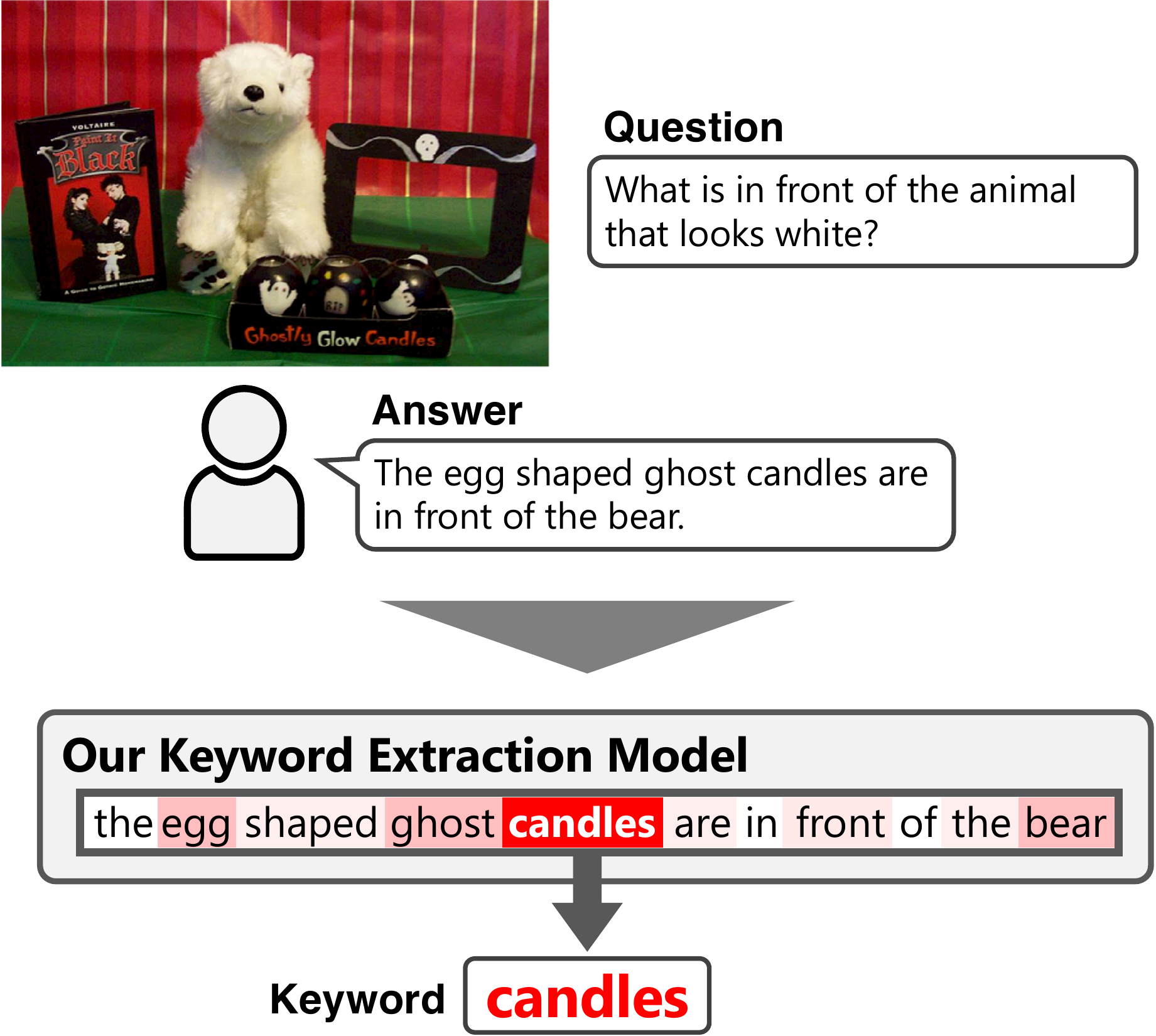}
\end{center}
\caption{
Example of the proposed task -- keyword extraction from full-sentence VQA.
Given an image, the question, and the full-sentence answer, the keyword extraction model extracts a keyword from the full-sentence answer.
In this example, the word ``candles'' is the most important part, answering the question ``What is in front of the animal that looks white?''.
Therefore, ``candles'' is considered as the keyword of the answer.
}
\end{figure}

The ultimate goal of the present work is to gain knowledge through VQA that can be easily transferred to other tasks, such as object class recognition and object detection.
Therefore, the knowledge (VQA answers) should be represented by a single word, such as a class label.
However, in real-world dialog, answers are rarely expressed by single words; rather, they are often expressed as complete sentences.
In fact, in VisDial v1.0~\citep{visdial}, a dataset of natural conversations about images that does not have a word limit for answers, the average length of answers is 6.5 words.
This is significantly longer than the average length of the answers in the VQA v2 dataset (1.2 words).

To bridge the gap between existing VQA research and real-world VQA, a challenging problem must be solved: identifying the word in the sentence that corresponds to the answer to the question.
It must also be considered that full-sentence answers provided by humans are likely to follow a variety of sentence structures.
Thus, the traditional approaches, such as rule-based approaches based on Part-of-Speech tagging or shallow parsing, require a great deal of work on defining rules in order to extract the keywords.
Our key challenge is to propose a novel keyword extraction method that leverages information from images and questions as clues, without the heavy work of annotating keywords or defining the rules.

This work handles the task of extracting a keyword when a full-sentence answer is obtained from VQA (\textbf{Full-sentence VQA}).
The simplest approach to this task is to construct a dataset containing full-sentence answers and keyword annotations, and then train a model based on this dataset in a supervised manner.
However, the cost of constructing a VQA dataset with full-sentence answers and keyword annotations is very high.
If a keyword extraction model can be trained on a dataset without keyword annotations, we can eliminate the high cost of collecting keyword annotations.

We propose an unsupervised keyword extraction model using a full-sentence VQA dataset which contains no keyword annotations.
Here, the principle is based on the intuition that the keyword is the most informative word in the full-sentence answer, and contains the information that is not included in the question (i.e., the concise answer).
Essentially, the full-sentence answer can be decomposed into two types of words: (1) the keyword information that is not included in the question, and (2) the information that is already included in the question.
For example, in the answer ``The egg shaped ghost candles are in front of the bear.'' to the question ``What is in front of the animal that looks white?'', the word ``candles'' is the keyword, while the remaining part ``The egg shaped ghost \textit{something} is in front of the bear'' is either information already included in the question or additional information about the keyword.
In this case, words like ``egg,'' ``ghost,'' and ``bear'' are also not in the question, making it difficult to find the keyword via naive methods, e.g., rule-based keyword extraction.
Our proposed model utilizes image features and question features to calculate the importance score for each word in the full-sentence answer.
Therefore, based on the contents of the image and the question, the model can accurately estimate which words in the full-sentence answer are important.
To the best of our knowledge, this is the first attempt at extracting a keyword from full-sentence VQA in an unsupervised manner.
\begin{figure*}[t]
\begin{center}
\includegraphics[width=0.8\linewidth]{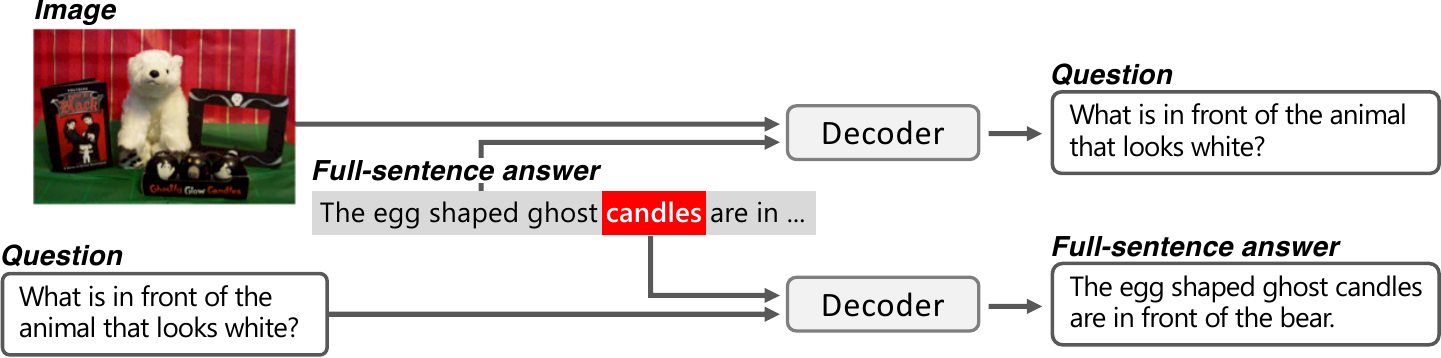}
\end{center}
\caption{
Illustration of the key concept.
In this example, the word ``candles'' is the keyword for the full-sentence answer, ``The egg shaped ghost candles are in front of the bear.''
We consider the keyword extraction task as the decomposition of the full-sentence answer into answer information and question information.
Therefore, if the keyword (i.e., the most informative word in the full-sentence answer) can be accurately extracted, the original full-sentence answer can be reconstructed from it.
Additionally, the question can be reconstructed from the decomposed question information in the full-sentence answer.
}
\label{fig:core}
\end{figure*}
The main contributions of this work are as follows:
(1) We propose a novel task of extracting keywords from full-sentence VQA with no keyword annotations.
(2) We designed a novel, unsupervised keyword extraction model by decomposing the full-sentence answer.
(3) We conducted experiments on two VQA datasets, and provided both qualitative and quantitative results that show the effectiveness of our model.

\section{Related Work}

\subsection{Unsupervised Keyword Extraction for Text}

Unsupervised keyword extraction methods can be broadly classified into two categories: graph-based methods and statistical methods.

Graph-based methods construct graphs from target documents by using co-occurrence between words~\citep{textrank,singlerank}.
These methods are only applicable to documents with a certain length, as they require the words in the document to co-occur multiple times.
The target document in this work is a full-sentence answer of VQA, whose average length is about 10 words.
Therefore, graph-based methods are not suitable here.

Statistical methods rely on statistics obtained from a document.
The most basic statistical method is TF-IDF~\citep{tfidf}, which calculates the term frequency and inverse document frequency and scores each word in the target document.
Recent work such as EmbedRank ~\citep{embedrank} have utilized word embeddings for the unsupervised keyword extraction.
EmbedRank calculates the cosine similarity between the candidate word (or phrase) embeddings and the sentence embeddings to retrieve the most representative word of the text.

\subsection{Visual Question Answering}
VQA is a well-known task that involves learning from image-related questions and answers.
The most popular VQA dataset is VQA v2~\citep{vqav2}, and much research has used this dataset for performance evaluations.
In VQA v2, the average number of words in an answer is only 1.2, and the variety of answers is relatively limited.

As stated in Section~\ref{intro}, in natural question answering by humans, the answers will be expressed as a sentence rather than a single word.
Some datasets that have both full-sentence answers and keyword annotations exist.

FSVQA~\citep{fsvqa} is a VQA dataset with answers in the form of full sentences.
In it, full-sentence answers are automatically generated by applying the numerous rule-based natural language processing patterns to the questions and single-word answers in the VQA v1 dataset~\citep{vqa}.

The recently proposed dataset, named GQA~\citep{gqa}, also contains automatically generated full-sentence answers.
This dataset is constructed on the Visual Genome~\citep{visualgenome}, which has rich and complex annotations about images, including dense captions, questions, and scene graphs.
The questions and answers (both single-word and full-sentence) in the GQA dataset are created from scene graph annotations of the images.

The full-sentence answers in both datasets described above are annotated automatically, i.e., not by humans.
Therefore, neither dataset has both full-sentence answers and manually annotated keywords.

\subsection{Attention}
The attention mechanism is a technique originally proposed in machine translation~\citep{nmt_attention}, aimed at focusing on the most important part of the input sequences for a task.
Since the method proposed herein utilizes an attention mechanism to calculate the importance score of the word in the full-sentence answer, some prior works on attention mechanisms are discussed.

In general, an attention mechanism essentially learns the mapping between a query and key-value pairs.
Transformer~\citep{transformer} is one of the most popular attention mechanisms for machine translation.
It enables machine translation without using recurrent neural networks, using a self-attention mechanism and feed-forward networks instead.

Another study uses an attention mechanism for weakly supervised keyword extraction~\citep{keyword-attention}.
They first trained a model for document classification and extracted the word to which the model pays ``attention'' to perform the classification.
This system requires additional annotations of document class labels to train the model, whereas we aim to extract keywords without any additional annotations.

\section{Model}

This section describes the proposed method in detail.

First, the principal concept of the model is shown in Figure~\ref{fig:core}.
To extract the keyword, we intend to obtain two features from the full-sentence answer, each representing the keyword information and the information derived from the question, respectively.
To ensure that these two features discriminatively include keyword information and question information, we intend to reconstruct the original questions and answers from the question features and keyword features, respectively.
Thus, if we successfully extract the keyword and the question information from the full-sentence answer, we can reconstruct original full-sentence answer and the question.
Essentially, given an image, its corresponding question, and full-sentence answer, our proposed model extracts the keyword of the answer by decomposing the keyword information and the question information in the answer.

\subsection{Overview}
\begin{figure*}[t]
\begin{center}
\includegraphics[width=0.8\linewidth]{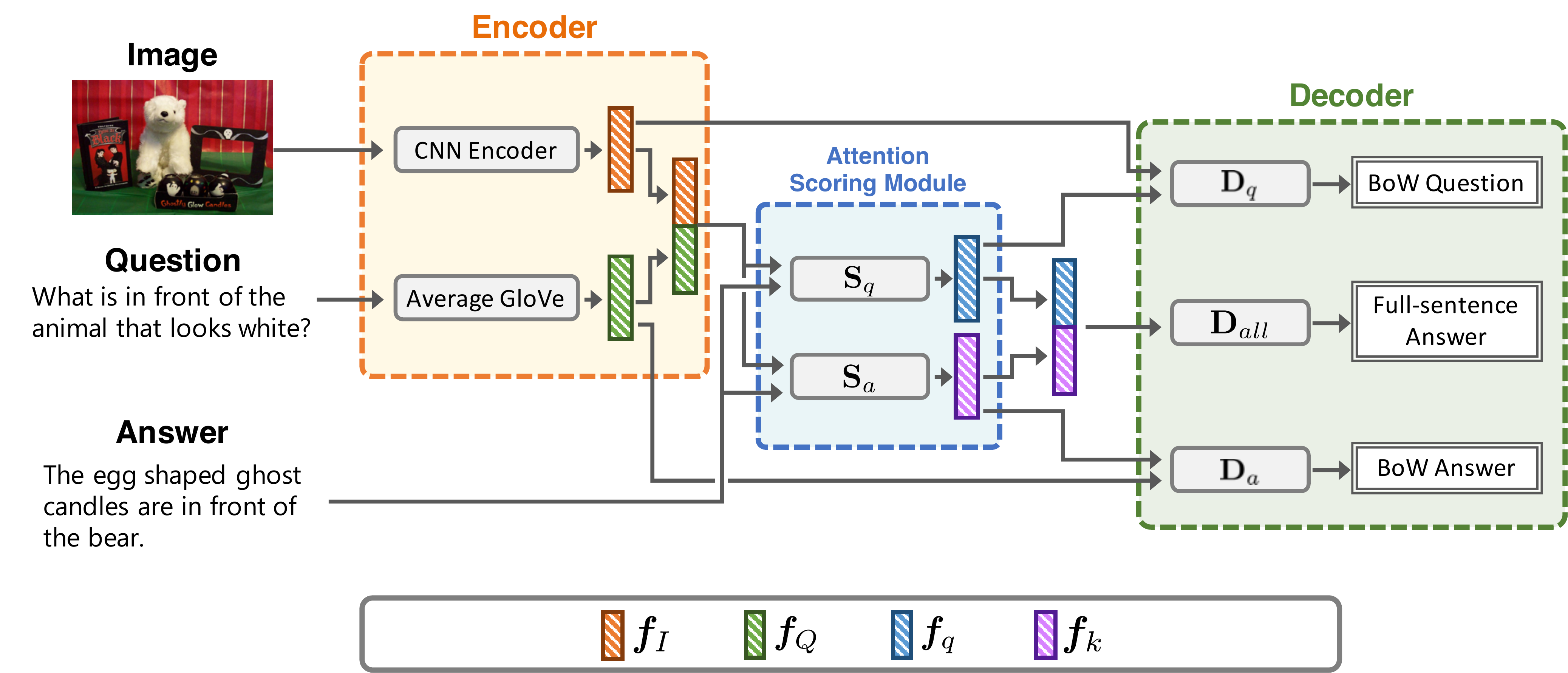}
\end{center}
\caption{
Overall pipeline of the model.
First, the Encoder Module extracts the image features $\bm{f}_I$ and the question features $\bm{f}_Q$ and integrates them into a joint feature $\bm{f}_j$.
Then, the Attention Scoring Modules $\mathbf{S}_a$ and $\mathbf{S}_q$ compute the attention weight and calculate the weighted sum of the word-embedding vectors of the full-sentence answer.
The output of $\mathbf{S}_a$ i.e., $\bm{f}_k$, is the keyword-aware feature of the full-sentence answer, and the output of $\mathbf{S}_q$ i.e., $\bm{f}_q$, is the question-aware feature.
$\mathbf{D}_{all}$ reconstructs the full-sentence answer from both $\bm{f}_k$ and $\bm{f}_q$.
$\mathbf{D}_a$ estimates the Bag-of-Words(BoW) feature of the full-sentence answer from $\bm{f}_k$ and $\bm{f}_Q$.
Additionally, $\mathbf{D}_q$ estimates the BoW feature of the question from $\bm{f}_q$ and $\bm{f}_I$.
}
\label{fig:model}
\end{figure*}

An overview of the model is shown in Figure ~\ref{fig:model}.
To realize decomposition-based keyword extraction, we designed a model which consists of the encoder $\mathbf{E}$, the attention scoring modules $\mathbf{S}_a$ and $\mathbf{S}_q$, and the decoder modules $\mathbf{D}_{all}$, $\mathbf{D}_a$, and $\mathbf{D}_q$.

An image $\bm{I}$ and the corresponding question $\bm{Q}$ and full-sentence answer $\bm{A}=\{w^{(a)}_1, \;w^{(a)}_2, \;\ldots,\;w^{(a)}_n\}$ are considered as the model input.
Here, $w^{(a)}_i$ represents the \textit{i}-th word in the full-sentence answer.

Given $\bm{I}$ and $\bm{Q}$, $\mathbf{E}$ extracts image and question features and integrates them into joint features $\bm{f}_j$, i.e., $\mathbf{E}(\bm{I}, \bm{Q}) = \bm{f}_j$.

Next, $\mathbf{S}_a$ and $\mathbf{S}_q$ use $\bm{f}_j$ and $\bm{A}$ as input and output the weight vectors $\bm{a}_k$ and $\bm{a}_q$.
Here, $\bm{a}_k = \{a_{1}^{(k)}, \;a_{2}^{(k)},\; \ldots,\;a_{n}^{(k)}\}$ and $\bm{a}_q = \{a_{1}^{(q)}, \;a_{2}^{(q)},\; \ldots,\;a_{n}^{(q)}\}$  for each word in $\bm{A}$.
We denote $a_i \in (0, 1)$ as the weight score of the $i$-th word in $\bm{A}$.

Then, we consider the keyword vector $\bm{f}_k$ as the embedding vector of the word with the highest weight score in $\bm{a}_k$.
Meanwhile, the question information vector $\bm{f}_q$ is considered as the weighted sum of the embedding vectors of $\bm{A}$ corresponding to the weight score $\bm{a}_q$.

Following this, $\mathbf{D}_{all}$ uses LSTM to reconstruct the original full-sentence answer using $\bm{f}_q$ and $\bm{f}_k$.
$\bm{f}_q$ and $\bm{f}_k$ are intended to represent the question information and the keyword vector of the full-sentence answer, respectively.
However, $\mathbf{D}_{all}$ only ensures that both features have the information of the full-sentence answer.
To separate them, we designed the additional decoders, $\mathbf{D}_{a}$ and $\mathbf{D}_{q}$.
The former reconstructs the BoW features of the answer using $\bm{f}_k$, while the latter reconstructs those of the question using $\bm{f}_{q}$ with auxiliary vectors.
The objective of this operation is to make $\bm{f}_k$ and $\bm{f}_{q}$ representative features for the full-sentence answer and the question, respectively.

The entire model is trained to minimize the disparity between the reconstructed sentences $\bm{A}_{recon}$ and the original full-sentence answers, as well as that between the BoW features of the full-sentence answers and the questions.

\subsection{Encoder}
The module $\mathbf{E}$ encodes the image $\bm{I}$ and the question $\bm{Q}$ and obtains the image feature $\bm{f}_I$, the question feature $\bm{f}_{Q}$, and the joint feature $\bm{f}_j$.
To generate $\bm{f}_I$, we use the image feature extracted from a deep CNN, which is pre-trained on a large-scale image recognition dataset.
For $\bm{f}_{Q}$, each word token was converted into a word embeddings and averaged.
Following this, $l_2$ normalization was performed on both features.
Finally, those features were concatenated to the joint feature $\bm{f}_j \in \mathbb{R}^{d_j}$, i.e., $\mathbf{E}(\bm{I},\,\bm{Q}) = \bm{f}_j = [\bm{f}_I; \,\bm{f}_{Q}]$, where $d_j$ is the dimension of the joint feature and $[;]$ indicates concatenation.
Note that we did not update the model parameters of $\mathbf{E}$ during training.

\subsection{Attention Scoring Module}

This module takes $\bm{f}_j$ as input and weights each words in the full-sentence answer.
We used two of these modules, $\mathbf{S}_a$ and $\mathbf{S}_q$.
$\mathbf{S}_a$ and $\mathbf{S}_q$ compute the weights based on the importance of a word for the full-sentence answer and that for the question, respectively.
$\mathbf{S}_a$ and $\mathbf{S}_q$ have a nearly identical structure.
Therefore, the details of $\mathbf{S}_a$ are presented first, following which the difference between $\mathbf{S}_a$ and $\mathbf{S}_q$ is described.

The weight scoring in these modules is based on the attention mechanism used in Transformer~\citep{transformer}.
First, each word in the full-sentence answer was encoded, and the full-sentence answer vector $\bm{f}_A = \{\bm{w}^{(a)}_1, \;\bm{w}^{(a)}_2, \;\ldots,\;\bm{w}^{(a)}_n\} \in \mathbb{R}^{d_e \times n}$ was created.
Here, $\bm{w}^{(a)}_i$ denotes the embedding vector of the i-th word, $n$ is the length of the full-sentence answer, and $d_e$ is the dimension of the word embedding vector.
To represent the word order, positional encoding was applied to $\bm{f}_A$.
Specifically, before feeding $\bm{f}_A$ into scoring modules, we add positional embedding vectors to $\bm{f}_A$, similar to those introduced in BERT~\citep{bert}.

We describe our attention mechanism as a mapping between Query and Key-Value pairs.
First, we calculate Query vector $\bm{Q} \in \mathbb{R}^{h}$, Key vector $\bm{K} \in \mathbb{R}^{h \times n}$, and Value vector $\bm{V} \in \mathbb{R}^{h \times n}$.

\begin{align}
\bm{Q} &= \textrm{FFN}_q(\bm{f}_j) \\
\bm{K} &= \textrm{FFN}_k(\bm{f}_A) \\
\bm{V} &= \textrm{FFN}_v(\bm{f}_A) = \{\bm{v}^{(a)}_1, \;\bm{v}^{(a)}_2, \;\ldots,\;\bm{v}^{(a)}_n\}
\end{align}
where $\textrm{FFN}_q$, $\textrm{FFN}_k$, $\textrm{FFN}_v$ are single-layer feed-forward neural networks.
Then, the attention weight vector $\bm{a}_k = \{a_1^{(k)},\,a_2^{(k)},\,\dots,\,a_n^{(k)}\} \in \mathbb{R}^{n}$, where $a_i^{(k)}$ is the weighted score of the \textit{i}-th word, is computed as the product of $\bm{Q}$ and $\bm{K}$, as shown below.
\begin{equation}
\bm{a}_k = \bm{K}^{\mathsf{T}} \bm{Q}
\end{equation}

Then, the word with the highest weighted score is chosen as the keyword of the full-sentence answer:
\begin{align}\label{argmax}
i^{(k)} &= \underset{i}{\textrm{argmax}}(a_i^{(k)}) \\
\bm{f}_k &= \bm{v}^{(a)}_{i^{(k)}}
\end{align}

However, the argmax operation is non-differentiable.
Therefore, we use an approximation of this operation by softmax with temperature.

\begin{align}
\bm{f}_k = \bm{V}\, {\textrm{softmax} } (\frac{\bm{a}_k}{\tau})
\end{align}
where $\tau$ is a temperature parameter, and as $\tau$ approaches 0, the output of the softmax function becomes a one-hot distribution.

$\mathbf{S}_q$ has the same structure as $\mathbf{S}_a$ up to the point of computing the attention weight vector $\bm{a}_q$.
For the keyword vector, we have the intention to focus on the specific word in the full-sentence answer.
Therefore, we use the softmax with temperature.
However, for the question vector, there is no need to focus on one word.
Therefore, the question vector is calculated as the weighted sum of the attention score:
\begin{align}
\bm{f}_q = \bm{V}\, {\textrm{softmax} } (\bm{a}_q)
\end{align}

Then, we applied single-layer feed-forward neural network, followed by layer normalization~\citep{layernorm} to the output of this module $\bm{f}_k, \bm{f}_q$.

\subsection{Decoder}
\paragraph{Entire Decoder}
In the entire decoder $\mathbf{D}_{all}$, the full-sentence is reconstructed from the output of the attention scoring modules $\bm{f}_k$ and $\bm{f}_q$, i.e., $\bm{A}_{recon} = \mathbf{D}_{all}(\bm{f}_k, \bm{f}_q)$, where $\bm{A}_{recon}$ denotes the reconstructed full-sentence answer.
We use an LSTM as the sentence generator.
As the input to the LSTM at each step $(x_t)$, $\bm{f}_k$ and $\bm{f}_q$ are concatenated to the output of the previous step as follows:
\begin{align}
x_0 &= \bm{W}_{x_0} [\bm{f}_k;\,\bm{f}_q] \\
x_t &= \bm{W}_{x} \,[\bm{f}_k;\,\bm{f}_q;\,\hat{s}_{t-1}]
\end{align}
where $\hat{s}_{t-1}$ is the output of the LSTM at the $t-1$ step, and  $\bm{W}_{x_0}$ and $\bm{W}_{x}$ are the learned parameters.

The objective of $\mathbf{D}_{all}$ is defined by the cross- entropy loss:
\begin{equation}
L_{all} = -\sum_{t=1}^{n} \log (p(\hat{s}_t = s^{(ans)}_t \;| \;s^{(ans)}_{1:t-1}))
\end{equation}
where $s^{(ans)}$ is the ground-truth full-sentence answer.

Further, word dropout~\citep{cvae}, a method of masking input words with a specific probability, is applied.
This forces the decoder to generate sentences based on the $\bm{f}_k$ and $\bm{f}_q$ rather than relying on the previous word.

\paragraph{Discriminative Decoders}
$\mathbf{D}_{all}$ attempts to reconstruct the full-sentence answer from $\bm{f}_k$ and $\bm{f}_q$.
Thus, $\mathbf{D}_{all}$ allows the feature vectors to contain the answer information.
However, the keyword and question information are intended to be represented by $\bm{f}_k$ and $\bm{f}_q$, respectively.
Therefore, we designed the discriminative decoders, $\mathbf{D}_a$ and $\mathbf{D}_q$, to generate $\bm{f}_k$ and $\bm{f}_q$, respectively, thus capturing the desired information separately.

$\mathbf{D}_a$ and $\mathbf{D}_q$ reconstruct the full-sentence answer and the question, respectively.
This reconstruction is performed with the target of the BoW features of the sentence, rather than the sentence itself.
This is because we intend to focus on the content of the sentence and not its sequential information.
Sentence reconstruction was also considered as an alternative, but this is difficult to train using LSTM.
The BoW feature $\bm{b} \in \mathbb{R}^{n_s}$ is represented as a vector whose \textit{i}-th elements is $N_i/L_s$, where $n_s$ is the vocabulary size, $N_i$ is the number of occurrences of the \textit{i}-th word, and $L_s$ is the number of the words in the sentence.

The input to these discriminative decoders consists not only of feature vectors, but also auxiliary vectors, the additional features that assist in reconstruction.
Specifically, the auxiliary vector for $\mathbf{D}_a$ is the average of the word embedding vectors in the question, $\bm{f}_{Q}$, and, for $\mathbf{D}_q$, the auxiliary vector is the image feature $\bm{f}_I$.

We build the decoder as the following fully-connected layers:
\begin{align}
\bm{y}_a &= \bm{W}_A[\bm{f}_k;\,\bm{f}_{Q}] + \bm{B}_A \\
\bm{y}_q &= \bm{W}_Q[\bm{f}_q;\,\bm{f}_I] + \bm{B}_Q
\end{align}

The loss function for the discriminative decoder is the cross-entropy loss between the ground-truth BoW features and the predicted BoW features:
\begin{align}
L_{a} &= -\sum_{i=1}^{n_a}\bm{b}_{a}[i]\, \log(\textrm{softmax}(\bm{y}_a[i])) \\
L_{q} &= -\sum_{i=1}^{n_q}\bm{b}_{q}[i]\, \log(\textrm{softmax}(\bm{y}_q[i]))
\end{align}
where $\bm{b}$ denotes the ground-truth of the BoW features, and $n_a$ and $n_q$ are the vocabulary sizes of the answer and the question, respectively.

\subsection{Full Objectives}
Finally, the overall objective function for the proposed model is written as

\begin{equation}
L = \lambda_{all} L_{all} + \lambda_{a} L_{a} + \lambda_{q} L_{q},
\end{equation}
where $\lambda_{all}$, $\lambda_{a}$, and $\lambda_{q}$ are hyper-parameters that balance each loss function.

\subsection{Implementation Details}
In the encoder $\mathbf{E}$, image features of size $2048 \times 14 \times 14$ were extracted from the pool-5 layer of the ResNet152~\citep{resnet}.
These were pre-trained on ImageNet, and global pooling was applied to obtain 2048-dimensional features.
To encode the question words, we used 300-dimensional GloVe embeddings~\citep{glove}.
These were pre-trained on the Wikipedia / Gigaword corpus\footnote{http://nlp.stanford.edu/projects/glove/}.

To convert each word in the full-sentence answer into $\bm{f}_A$, the embedding matrix in the attention scoring module was initialized with the pre-trained GloVe embeddings.
The temperature parameter $\tau$ is gradually annealed using the schedule $\tau_{i} = \max (\tau_{0} \;e^{-ri}, \tau_{min})$, where $i$ is the overall training iteration, and other parameters are set as $\tau_{0}=0.5, \,r=3.0\times10^{-5},\,\tau_{\min}=0.1$.
The LSTM in the $\mathbf{D}_{all}$ has a hidden state of 1024 dimensions.
The word dropout rate was set to 0.25.

We used the Adam~\citep{adam} optimizer to train the model, which has an initial learning rate of $1.0\times10^{-3}$.

\section{Experimental Setup}
\subsection{Dataset}
\begin{table}[t]
\centering
\begin{tabular}{@{}lcccc@{}}
\toprule
\textbf{Dataset} &  & \textbf{Size} & \textbf{Answer length}  \\ \midrule
\multirow{2}{*}{GQA} & train & 943,000 & 6.69 \\
& val & 132,062 & 6.70 &  \\ \midrule
\multirow{2}{*}{FSVQA} & train & 139,038 & 6.11  \\
& val & 68,265 & 6.07 &  \\ \midrule
\multirow{2}{*}{VQA v2} & train & 443,757 & 1.16  \\
& val & 214,354 & 1.16 &  \\ \bottomrule
\end{tabular}\vspace{2mm}
\vspace{-4mm}

\caption{\label{tab:dataset}
Basic statistics of the dataset we used.
Note that these FSVQA dataset values were taken after pre-processing.
Although VQA v2 was not used in this work, it is included in this table for reference.
}
\end{table}

We conducted experiments on two datasets: GQA and FSVQA.
In Table~\ref{tab:dataset}, we present the basic statistics of both datasets.

\noindent\textbf{GQA}~
GQA~\citep{gqa} contains 22M questions and answers.
The questions and answers are automatically generated from image scene graphs, and the answers include both the single-word answers and the full-sentence answers.
The questions and answers in GQA have unbalanced answer distributions.
Therefore, we used a balanced version of this dataset, which is down-sampled from the original dataset and contains 1.7M questions.
As pre-processing, we removed the periods, commas, and question marks.

\noindent\textbf{FSVQA}
FSVQA~\citep{fsvqa} contains 370K questions and full-sentence answers.
This dataset was built by applying rule-based processing to the VQA v1 dataset~\citep{vqa}, and captions in the MSCOCO dataset~\citep{mscoco}, to obtain the full-sentence answers.
There are ten annotations (i.e., single-word answers) per question in the VQA v1 dataset.
Of these, the annotations with the highest frequency is chosen to create full-sentence answers.
If all the frequencies are equal, an annotation is chosen at random.
Since the authors do not provide the mapping between single-word answers and full-sentence answers, we considered the annotations with the highest frequency as the single-word answers matching the full-sentence answers.
Questions for which the highest frequency annotation cannot be determined were filtered out.
Following this process, we obtained 139,038 questions for the training set, and 68,265 questions for the validation set.

\subsection{Settings}
The model performance was determined based on the keyword accuracy and the Mean Rank.
Mean Rank is the average rank of the correct keyword when sorting each word in order of the importance score.
Mean Rank is formulated as:
\begin{equation}
\textrm{Mean Rank} = \frac{1}{N} \sum_{i} rank_{i}.
\end{equation}
Here, $rank_{i}$ is the number representing the keyword rank when the words in the $i$-th answer sentence are arranged in order of the importance (TF-IDF score or attention score, i.e., $a^{(k)}_i$ in Eqn.~\ref{argmax}), and $N$ is the size of the overall samples.

We ran experiments with the various existing unsupervised keyword extraction methods for the comparison: (1) TF-IDF~\citep{tfidf}, (2) YAKE~\citep{yake}, and (3) EmbedRank~\citep{embedrank}.
Since YAKE removes the words with less than three characters as preprocessing, the Mean Rank cannot be calculated under the same conditions as other methods.
Therefore, the Mean Rank of YAKE is not shown.
We also conducted an ablation study to show the importance of $\mathbf{D}_a$ and $\mathbf{D}_q$.
In addition, we changed the reconstruction method from BoW estimation to the original sentence generation using LSTM.

\begin{table*}[t]
\centering
\begin{tabular}{@{}lccccc@{}}
\toprule
\multicolumn{1}{c}{} &\multicolumn{2}{c}{\rule{0pt}{2.3ex}\textbf{GQA}} & & \multicolumn{2}{c}{\textbf{FSVQA}}\\
\textbf{Model}   \rule{0pt}{2.3ex} & \textbf{Accuracy} ($\uparrow$) & \textbf{Mean Rank} ($\downarrow$) && \textbf{Accuracy} ($\uparrow$) & \textbf{Mean Rank} ($\downarrow$)    \\\midrule[0.05em]
TF-IDF   \rule{0pt}{2.0ex} &  0.275 & 2.86  &&   0.278   &   3.22      \\
YAKE  \rule{0pt}{2.0ex} &  0.269   & --  &&   0.107   &   --      \\
EmbedRank \rule{0pt}{2.0ex} &  0.306   & 2.15  &&   0.302   &   \textbf{2.32}      \\
Ours  \rule{0pt}{2.0ex} &\textbf{0.429 $\pm$ 0.03} &\textbf{2.04 $\pm$ 0.10} && \textbf{0.351 $\pm$ 0.04} & 2.38 $\pm$ 0.14      \\
Ours w/o $\mathbf{D}_{q}$ \rule{0pt}{2.0ex}     & 0.318 $\pm$ 0.02    & 2.44 $\pm$ 0.04        &&    0.298 $\pm$ 0.03       &        2.67 $\pm$ 0.05      \\
Ours w/o $\mathbf{D}_{a},\,\mathbf{D}_{q}$\rule{0pt}{2.0ex}  & 0.350  $\pm$ 0.06   & 3.01  $\pm$ 0.49  &&     0.347 $\pm$ 0.05      &         2.49 $\pm$ 0.21     \\
Ours (LSTM $\mathbf{D}_{a},\,\mathbf{D}_{q}$) \rule{0pt}{2.0ex} & 0.329  $\pm$ 0.01   & 2.36  $\pm$ 0.04  &&      0.347 $\pm$ 0.01     &      3.35 $\pm$ 0.11        \\\bottomrule
\end{tabular}\vspace{2mm}
\vspace{-2mm}
\caption{\label{tab:result_all}
Keyword extraction performanc on GQA and FSVQA.
Higher accuracies and lower Mean Ranks are desirable.
We conducted experiments three times using the proposed method.
Note that comparison methods are deterministic algorithms, experiments with them were conducted only once.
}
\end{table*}

\begin{figure*}
   \centering
   \includegraphics[width=0.85\linewidth]{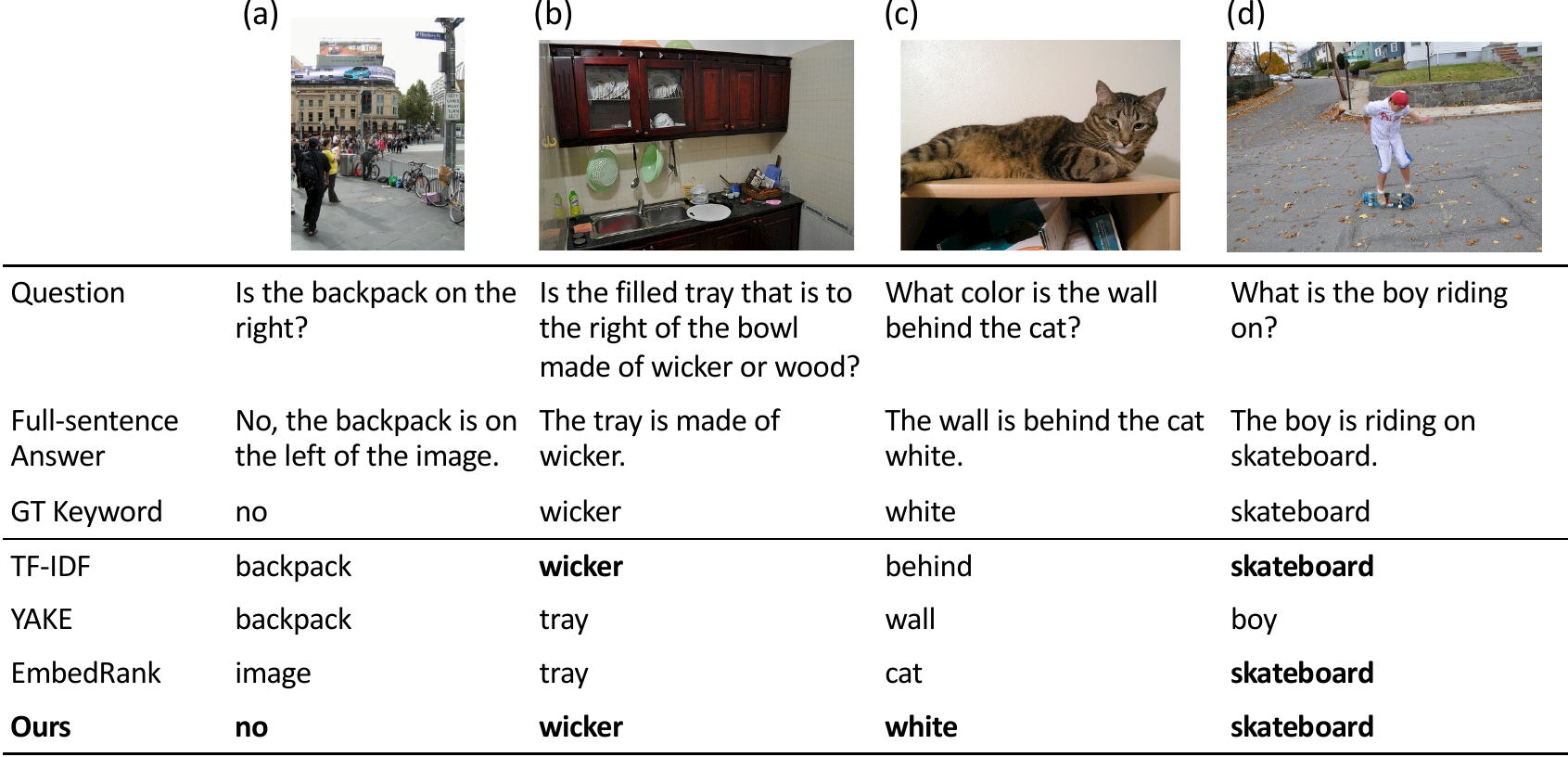}
   \caption{Examples of the keyword extraction results in the GQA dataset (a, b) and the FSVQA dataset (c, d).}
   \label{fig:gqa_quality}
\end{figure*}

\section{Experimental Results}

The experimental results are shown in Table~\ref{tab:result_all}.
Also, we provide the accuracy per question types in Appendix ~\ref{per_question} for further analysis.
The proposed model, which used BoW estimation in $\mathbf{D}_{a}$ and $\mathbf{D}_{q}$, achieves superior performance on almost all metrics and datasets except for the Mean Rank of FSVQA.
As can be seen in the results of the ablation study, this superior performance is achieved even without $\mathbf{D}_{a}$ and $\mathbf{D}_q$, which demonstrates the effectiveness of the proposed reconstruction-based method.
When using LSTM in $\mathbf{D}_{a}$ and $\mathbf{D}_q$, the accuracy and mean rank worsens as compared to those of the proposed model, which reconstructs the BoW in those modules.
This is considered to be because sentence reconstruction with LSTM requires management of the sequential information of the sentence, which is more complex than BoW estimation.
Since we intended to focus on the contents of the sentence, the BoW is more suitable for these modules.

We provide some examples in Figure ~\ref{fig:gqa_quality}.
The examples on the left and right are from GQA and FSVQA, respectively.
Since the statistical methods such as TF-IDF tend to choose rarer words as keywords, they are likely to fail if the keyword is a common word (Figure ~\ref{fig:gqa_quality} (a), (c)).
On the other hand, the model proposed herein can accurately extract keywords even in such cases.

\section{Conclusion}
In this paper, we proposed the novel task of unsupervised keyword extraction from full-sentence VQA.
A novel model was designed to handle this task based on information decomposition of full-sentence answers and the reconstruction of questions and answers.
Both qualitative and quantitative experiments show that our model successfully extracts the keyword of the full-sentence answer with no keyword supervision.

In future work, the extracted keywords will be utilized in other tasks, such as VQA, object classification, or object detection.
This work could also be combined with recent works on VQG~\citep{vqg_unknown,vqg_caption}.
In these works, the system generates questions to acquire information from humans.
However, they assume that the answers are obtained as single words, which will pose a problem when applying it to the real-world question answering.
By combining these studies with our research, an intelligent system can ask humans about unseen objects and learn new knowledge from the answer, even if the answer consists of more than a single word.

\noindent\textbf{Acknowledgement}
This work was partially supported by JST CREST Grant Number JPMJCR1403, and partially supported by JSPS KAKENHI Grant Number JP19H01115 and JP20H05556.
We would like to thank Yang Li, Sho Maeoki, Sho Inayoshi, and Antonio Tejero-de-Pablos for helpful discussions.

\bibliographystyle{acl_natbib}
\bibliography{emnlp2020}

\begin{thebibliography}{24}
\expandafter\ifx\csname natexlab\endcsname\relax\def\natexlab#1{#1}\fi

\bibitem[{Antol et~al.(2015)Antol, Agrawal, Lu, Mitchell, Batra,
  Lawrence~Zitnick, and Parikh}]{vqa}
Stanislaw Antol, Aishwarya Agrawal, Jiasen Lu, Margaret Mitchell, Dhruv Batra,
  C.~Lawrence~Zitnick, and Devi Parikh. 2015.
\newblock Vqa: Visual question answering.
\newblock In \emph{ICCV}.

\bibitem[{Ba et~al.(2016)Ba, Kiros, and Hinton}]{layernorm}
Jimmy~Lei Ba, Jamie~Ryan Kiros, and Geoffrey~E Hinton. 2016.
\newblock Layer normalization.
\newblock \emph{arXiv preprint arXiv:1607.06450}.

\bibitem[{Bahdanau et~al.(2015)Bahdanau, Cho, and Bengio}]{nmt_attention}
Dzmitry Bahdanau, Kyunghyun Cho, and Yoshua Bengio. 2015.
\newblock {Neural Machine Translation by Jointly Learning to Align and
  Translate}.
\newblock In \emph{ICLR}.

\bibitem[{Bennani-Smires et~al.(2018)Bennani-Smires, Musat, Hossmann,
  Baeriswyl, and Jaggi}]{embedrank}
Kamil Bennani-Smires, Claudiu Musat, Andreea Hossmann, Michael Baeriswyl, and
  Martin Jaggi. 2018.
\newblock Simple unsupervised keyphrase extraction using sentence embeddings.
\newblock In \emph{CoNLL}.

\bibitem[{Bowman et~al.(2016)Bowman, Vilnis, Vinyals, Dai, Jozefowicz, and
  Bengio}]{cvae}
Samuel~R. Bowman, Luke Vilnis, Oriol Vinyals, Andrew Dai, Rafal Jozefowicz, and
  Samy Bengio. 2016.
\newblock Generating sentences from a continuous space.
\newblock In \emph{CoNLL}.

\bibitem[{Campos et~al.(2020)Campos, Mangaravite, Pasquali, Jorge, Nunes, and
  Jatowt}]{yake}
Ricardo Campos, V^^c3^^adtor Mangaravite, Arian Pasquali, Al^^c3^^adpio Jorge,
  C^^c3^^a9lia Nunes, and Adam Jatowt. 2020.
\newblock Yake! keyword extraction from single documents using multiple local
  features.
\newblock \emph{Information Sciences}.

\bibitem[{Das et~al.(2017)Das, Kottur, Gupta, Singh, Yadav, Moura, Parikh, and
  Batra}]{visdial}
Abhishek Das, Satwik Kottur, Khushi Gupta, Avi Singh, Deshraj Yadav, Jose M.~F.
  Moura, Devi Parikh, and Dhruv Batra. 2017.
\newblock Visual dialog.
\newblock In \emph{CVPR}.

\bibitem[{Devlin et~al.(2019)Devlin, Chang, Lee, and Toutanova}]{bert}
Jacob Devlin, Ming-Wei Chang, Kenton Lee, and Kristina Toutanova. 2019.
\newblock {BERT}: Pre-training of deep bidirectional transformers for language
  understanding.
\newblock In \emph{NAACL-HLT}.

\bibitem[{Goyal et~al.(2017)Goyal, Khot, Summers{-}Stay, Batra, and
  Parikh}]{vqav2}
Yash Goyal, Tejas Khot, Douglas Summers{-}Stay, Dhruv Batra, and Devi Parikh.
  2017.
\newblock Making the {V} in {VQA} matter: Elevating the role of image
  understanding in {V}isual {Q}uestion {A}nswering.
\newblock In \emph{CVPR}.

\bibitem[{He et~al.(2016)He, Zhang, Ren, and Sun}]{resnet}
Kaiming He, Xiangyu Zhang, Shaoqing Ren, and Jian Sun. 2016.
\newblock Deep residual learning for image recognition.
\newblock In \emph{CVPR}.

\bibitem[{Hudson and Manning(2019)}]{gqa}
Drew~A. Hudson and Christopher~D. Manning. 2019.
\newblock Gqa: A new dataset for real-world visual reasoning and compositional
  question answering.
\newblock In \emph{CVPR}.

\bibitem[{Kingma and Ba(2015)}]{adam}
Diederik~P Kingma and Jimmy Ba. 2015.
\newblock Adam: A method for stochastic optimization.
\newblock In \emph{ICLR}.

\bibitem[{Krishna et~al.(2017)Krishna, Zhu, Groth, Johnson, Hata, Kravitz,
  Chen, Kalantidis, Li, Shamma et~al.}]{visualgenome}
Ranjay Krishna, Yuke Zhu, Oliver Groth, Justin Johnson, Kenji Hata, Joshua
  Kravitz, Stephanie Chen, Yannis Kalantidis, Li-Jia Li, David~A Shamma, et~al.
  2017.
\newblock Visual genome: Connecting language and vision using crowdsourced
  dense image annotations.
\newblock \emph{IJCV}.

\bibitem[{Lin et~al.(2014)Lin, Maire, Belongie, Hays, Perona, Ramanan,
  Doll{\'a}r, and Zitnick}]{mscoco}
Tsung-Yi Lin, Michael Maire, Serge Belongie, James Hays, Pietro Perona, Deva
  Ramanan, Piotr Doll{\'a}r, and C~Lawrence Zitnick. 2014.
\newblock Microsoft coco: Common objects in context.
\newblock In \emph{ECCV}.

\bibitem[{Mihalcea and Tarau(2004)}]{textrank}
Rada Mihalcea and Paul Tarau. 2004.
\newblock {T}ext{R}ank: Bringing order into text.
\newblock In \emph{EMNLP}.

\bibitem[{Misra et~al.(2018)Misra, Girshick, Fergus, Hebert, Gupta, and van~der
  Maaten}]{lba}
Ishan Misra, Ross Girshick, Rob Fergus, Martial Hebert, Abhinav Gupta, and
  Laurens van~der Maaten. 2018.
\newblock Learning by asking questions.
\newblock In \emph{CVPR}.

\bibitem[{Pennington et~al.(2014)Pennington, Socher, and Manning}]{glove}
Jeffrey Pennington, Richard Socher, and Christopher Manning. 2014.
\newblock {G}love: Global vectors for word representation.
\newblock In \emph{EMNLP}.

\bibitem[{Ramos(2003)}]{tfidf}
Juan Ramos. 2003.
\newblock {Using TF-IDF to Determine Word Relevance in Document Queries}.
\newblock In \emph{the first instructional conference on machine learning}.

\bibitem[{Shen et~al.(2019)Shen, Kar, and Fidler}]{vqg_caption}
Tingke Shen, Amlan Kar, and Sanja Fidler. 2019.
\newblock Learning to caption images through a lifetime by asking questions.
\newblock In \emph{ICCV}.

\bibitem[{Shin et~al.(2016)Shin, Ushiku, and Harada}]{fsvqa}
Andrew Shin, Yoshitaka Ushiku, and Tatsuya Harada. 2016.
\newblock The color of the cat is gray: 1 million full-sentences visual
  question answering (fsvqa).
\newblock \emph{arXiv preprint arXiv:1609.06657}.

\bibitem[{Uehara et~al.(2018)Uehara, Tejero-De-Pablos, Ushiku, and
  Harada}]{vqg_unknown}
Kohei Uehara, Antonio Tejero-De-Pablos, Yoshitaka Ushiku, and Tatsuya Harada.
  2018.
\newblock Visual question generation for class acquisition of unknown objects.
\newblock In \emph{ECCV}.

\bibitem[{Vaswani et~al.(2017)Vaswani, Shazeer, Parmar, Uszkoreit, Jones,
  Gomez, Kaiser, and Polosukhin}]{transformer}
Ashish Vaswani, Noam Shazeer, Niki Parmar, Jakob Uszkoreit, Llion Jones,
  Aidan~N Gomez, {\L}ukasz Kaiser, and Illia Polosukhin. 2017.
\newblock Attention is all you need.
\newblock In \emph{NeurIPS}.

\bibitem[{Wan and Xiao()}]{singlerank}
Xiaojun Wan and Jianguo Xiao.
\newblock Single document keyphrase extraction using neighborhood knowledge.

\bibitem[{Wu et~al.(2018)Wu, Du, and Guo}]{keyword-attention}
Xing Wu, Zhikang Du, and Yike Guo. 2018.
\newblock A visual attention-based keyword extraction for document
  classification.
\newblock \emph{Multimedia Tools and Applications}, 77(19):25355--25367.

\end{thebibliography}

\newpage
\appendix
\counterwithin{figure}{section}
\counterwithin{table}{section}

\section{Accuracy per Question Types}\label{per_question}

\begin{table*}[!htb]
\centering
\begin{tabular}[t]{c}
\begin{tabular}[t]{@{}lccccccccc@{}}
\toprule
\multirow{2}{*}{GQA} &\multicolumn{1}{l}{} &\multicolumn{2}{c}{Ours} &\multicolumn{2}{c}{TF-IDF} &\multicolumn{2}{c}{YAKE} &\multicolumn{2}{c}{EmbedRank}\\ \cmidrule(l){3-10}
& num. & Acc. &\begin{tabular}[c]{@{}c@{}}Mean\\ Rank\end{tabular} & Acc. &\begin{tabular}[c]{@{}c@{}}Mean\\ Rank\end{tabular} & Acc. &\begin{tabular}[c]{@{}c@{}}Mean\\ Rank\end{tabular} & Acc. &\begin{tabular}[c]{@{}c@{}}Mean\\ Rank\end{tabular} \\ \midrule

Is the & 22,507 & \textbf{{0.437}} & \textbf{{1.898}}& 0.058 & 3.819& 0.326 & --& 0.104 & 2.880 \\
What is & 20,435 & 0.544 & 1.444& \textbf{{0.581}} & \textbf{{1.295}}& 0.237 & --& 0.495 & 1.353 \\
Are there & 13,004 & 0.276 & 3.217& 0.000 & 4.859& 0.000 & --& \textbf{{0.395}} & \textbf{{2.020}} \\
Who is & 6,452 & 0.366 & \textbf{{1.773}}& 0.184 & 2.158& \textbf{{0.940}} & --& 0.080 & 2.541 \\
Is there & 5,270 & 0.269 & 2.971& 0.000 & 4.792& 0.000 & --& \textbf{{0.314}} & \textbf{{2.272}} \\
Does the & 5,236 & \textbf{{0.393}} & \textbf{{2.108}}& 0.031 & 3.812& 0.002 & --& 0.126 & 2.502 \\
On which & 5,121 & 0.369 & \textbf{{1.701}}& 0.001 & 2.074& \textbf{{0.997}} & --& 0.045 & 3.060 \\
Do you & 4,976 & 0.291 & 2.936& 0.005 & 4.770& 0.000 & --& \textbf{{0.338}} & \textbf{{2.118}} \\
Which kind & 4,410 & \textbf{{0.744}} & \textbf{{1.129}}& 0.720 & 1.143& 0.033 & --& 0.471 & 1.370 \\
What kind & 4,148 & \textbf{{0.719}} & \textbf{{1.171}}& 0.704 & 1.175& 0.060 & --& 0.476 & 1.358 \\

\bottomrule
\end{tabular} \vspace{2mm}
\\
\begin{tabular}{@{}lccccccccc@{}}
\toprule
\multirow{2}{*}{FSVQA} &\multicolumn{1}{l}{} &\multicolumn{2}{c}{Ours} &\multicolumn{2}{c}{TF-IDF} &\multicolumn{2}{c}{YAKE} &\multicolumn{2}{c}{EmbedRank}\\ \cmidrule(l){3-10}
& num. & Acc. &\begin{tabular}[c]{@{}c@{}}Mean\\ Rank\end{tabular} & Acc. &\begin{tabular}[c]{@{}c@{}}Mean\\ Rank\end{tabular} & Acc. &\begin{tabular}[c]{@{}c@{}}Mean\\ Rank\end{tabular} & Acc. &\begin{tabular}[c]{@{}c@{}}Mean\\ Rank\end{tabular} \\ \midrule

How many & 11,592 & \textbf{{0.271}} & \textbf{{2.579}}& 0.075 & 2.926& 0.003 & --& 0.004 & 3.930 \\
What is & 10,394 & 0.343 & 2.564& 0.542 & 1.724& 0.099 & --& \textbf{{0.551}} & \textbf{{1.698}} \\
What color & 10,237 & \textbf{{0.678}} & \textbf{{1.393}}& 0.073 & 2.385& 0.097 & --& 0.110 & 2.133 \\
Is the & 5,596 & \textbf{{0.336}} & \textbf{{2.312}}& 0.087 & 4.788& 0.046 & --& 0.163 & 2.452 \\
Is this & 4,475 & \textbf{{0.346}} & 2.172& 0.075 & 5.662& 0.074 & --& 0.279 & \textbf{{2.102}} \\
What are & 2,045 & 0.273 & 2.568& \textbf{{0.603}} & \textbf{{1.577}}& 0.007 & --& 0.602 & 1.587 \\
What kind & 1,695 & 0.245 & 2.393& \textbf{{0.889}} & \textbf{{1.160}}& 0.588 & --& 0.810 & 1.232 \\
Are the & 1,473 & \textbf{{0.291}} & \textbf{{2.367}}& 0.075 & 5.720& 0.025 & --& 0.131 & 2.499 \\
What type & 1,113 & 0.232 & 2.431& \textbf{{0.864}} & \textbf{{1.190}}& 0.543 & --& 0.812 & 1.226 \\
Where is & 1,106 & 0.500 & 2.042& \textbf{{0.568}} & \textbf{{1.626}}& 0.062 & --& 0.397 & 1.814 \\

\bottomrule
\end{tabular}
\end{tabular}
\vspace{1mm}
\caption{
\label{tab:per_question}
Performance per question types.
The upper and lower tables show the results of GQA and FSVQA, respectively.
}
\end{table*}

For further analysis, we report the performance of the model for each question type.
All questions are categorized based on the first two words, and the top 10 frequent question types are provided as per this categorization.
The results are shown in Table~\ref{tab:per_question}.

For most question types, our method shows higher accuracy and lower mean rank than the methods used for comparison.
The performance of the statistical methods drops to almost zero for some question types.
For example, the performance of the statistical methods for some question types where the answer is a common word, such as ``Is the'' or ``Are there,'' is very poor.

\section{Qualitative Results}

\begin{figure*}[!htbp]
   \centering
   \includegraphics[width=1.0\linewidth]{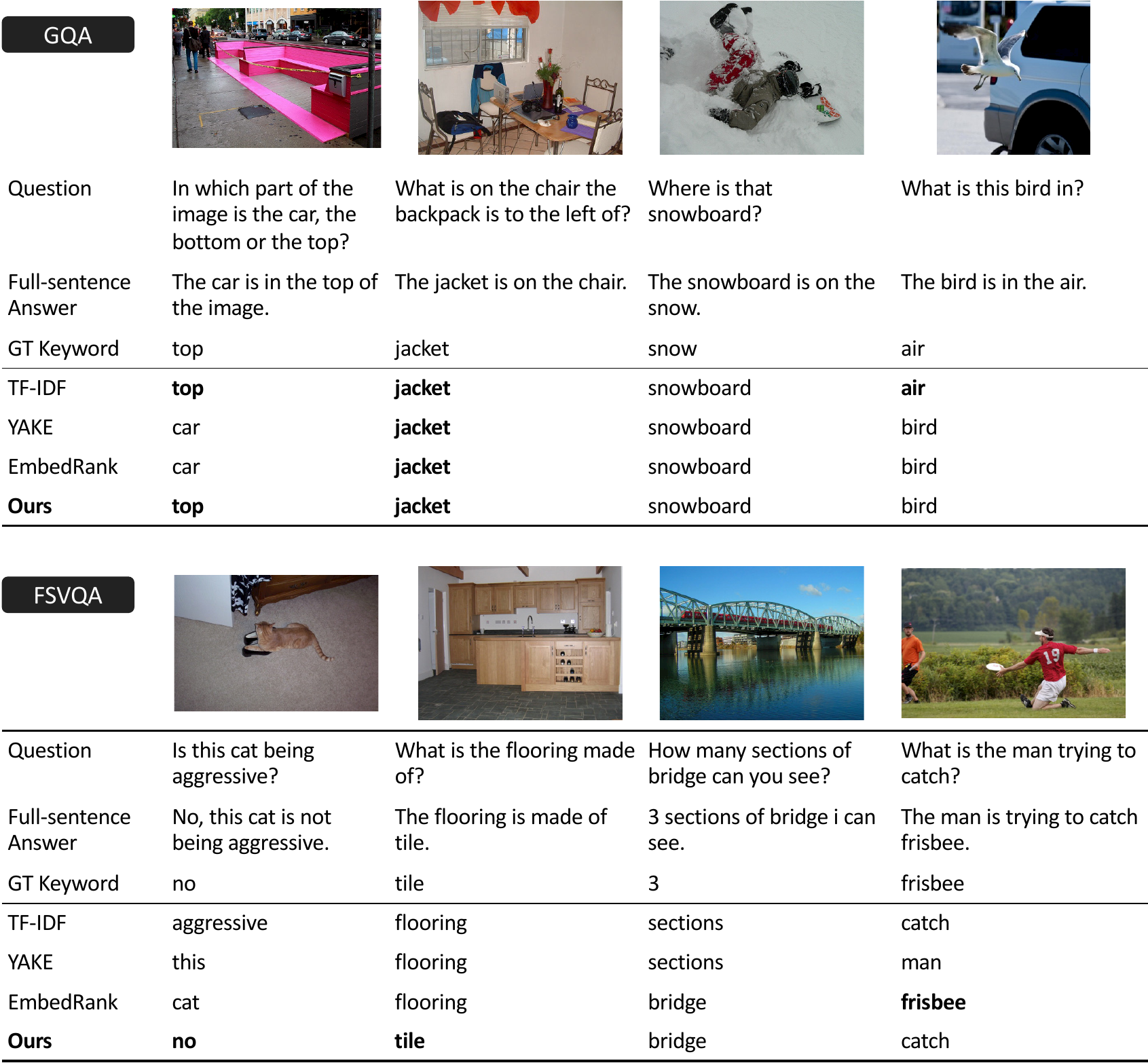}
   \caption{
   Examples of keyword extraction results using the GQA and FSVQA datasets.
   The upper and lower tables show the results of GQA and FSVQA, respectively.
   }
   \label{fig:gqa_supp}
\end{figure*}

Additional examples of qualitative results from applying the proposed model and comparison methods to the GQA and FSVQA datasets are shown in Figure~\ref{fig:gqa_supp}.

\clearpage

\end{document}